\documentclass[sigconf]{acmart}
\usepackage{bm}
\usepackage{multirow}
\usepackage[normalem]{ulem}

\newcommand{\etal}{\textit{et al}.}

\AtBeginDocument{%
  \providecommand\BibTeX{{%
    \normalfont B\kern-0.5em{\scshape i\kern-0.25em b}\kern-0.8em\TeX}}}

\copyrightyear{2020} 
\acmYear{2020} 
\setcopyright{acmlicensed}
\acmConference[MM '20]{Proceedings of the 28th ACM International Conference on Multimedia}{October 12--16, 2020}{Seattle, WA, USA}
\acmBooktitle{Proceedings of the 28th ACM International Conference on Multimedia (MM '20), October 12--16, 2020, Seattle, WA, USA}
\acmPrice{15.00}
\acmDOI{10.1145/3394171.xxxxxxx}
\acmISBN{978-1-4503-7988-5/20/10}

\begin{document}

\title{Neutral Face Game Character Auto-Creation via PokerFace-GAN}


\author{Tianyang Shi$^{1}$, Zhengxia Zou$^{2}$, Xinhui Song$^{1}$, Zheng Song$^{1}$}
\author{Changjian Gu$^{1}$, Changjie Fan$^{1}$, Yi Yuan$^{1*}$}
\affiliation{\institution{$^1$NetEase Fuxi AI Lab ($^*$Corresponding author)}}
\affiliation{\institution{$^2$University of Michigan, Ann Arbor}}
\email{[shitianyang, songxinhui, songzheng, guchangjian, fanchangjie, yuanyi]@corp.netease.com, zzhengxi@umich.edu}

\renewcommand{\shortauthors}{Shi and Zou, et al.}


\begin{abstract}
Game character customization is one of the core features of many recent Role-Playing Games (RPGs), where players can edit the appearance of their in-game characters with their preferences. This paper studies the problem of automatically creating in-game characters with a single photo. In recent literature on this topic, neural networks are introduced to make game engine differentiable and the self-supervised learning is used to predict facial customization parameters. However, in previous methods, the expression parameters and facial identity parameters are highly coupled with each other, making it difficult to model the intrinsic facial features of the character. Besides, the neural network based renderer used in previous methods is also difficult to be extended to multi-view rendering cases. In this paper, considering the above problems, we propose a novel method named ``PokerFace-GAN'' for neutral face game character auto-creation. We first build a differentiable character renderer which is more flexible than the previous methods in multi-view rendering cases. We then take advantage of the adversarial training to effectively disentangle the expression parameters from the identity parameters and thus generate player-preferred neutral face (expression-less) characters. Since all components of our method are differentiable, our method can be easily trained under a multi-task self-supervised learning paradigm. Experiment results show that our method can generate vivid neutral face game characters that are highly similar to the input photos. The effectiveness of our method is verified by comparison results and ablation studies.
\end{abstract}




\maketitle

\begin{figure}[h]
  \includegraphics[width=0.9\linewidth]{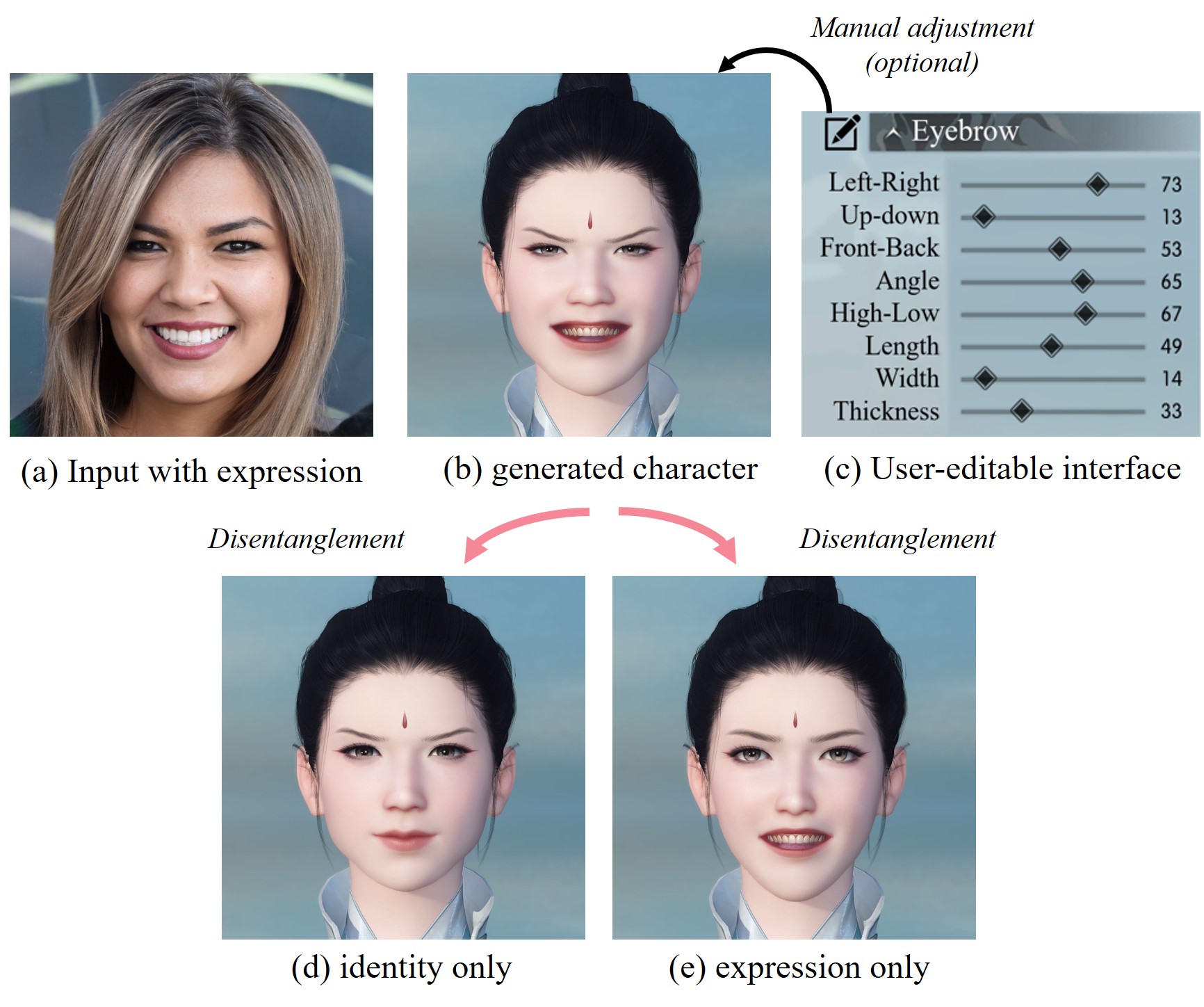}
  \caption{We propose a new method for neutral face game character auto-creation based on a single facial photo. Our method can effectively separate the character's expression and identity, which is rarely studied by previous works~\cite{Shi_2019_ICCV,shi2020fast}.}
  \Description{overview}
  \label{fig:cover}
\end{figure}

\section{Introduction}

Game characters are the core of role-playing games (RPGs). Recent RPGs commonly feature facial customization systems where players are allowed to customize their in-game characters according to their preference rather than using predefined templates. To improve the immersion of players, the facial customization systems have been remarkably developed and are becoming more and more sophisticated. 

As a result, players have to spend several hours manually adjusting hundreds of parameters in order to create the desired face (e.g.\ a movie star or the players themselves). Recently, the game character auto-creation is becoming an emerging research direction in computer vision and graphics. Many games (e.g. ``Justice\footnote{http://n.163.com}'' and ``NBA 2K20\footnote{https://nba.2k.com}'') introduced AI for creating 3D game characters based on single or multiple 2D photos. As a typical multimedia application, these AIs are popular with players since they can provide rich interest and interactivity.

Currently, there are two groups of methods for the auto-creation of 3D faces, where the first group is the 3D Morphable Model (3DMM) based methods~\cite{blanz1999morphable,BFM2009,Booth_2017_CVPR,genova2018unsupervised,gecer2019ganfit} and the second group is the Face-to-Parameter translation (F2P)~\cite{Shi_2019_ICCV,shi2020fast}. The 3DMM-based methods have developed significantly over the past two decades. This group of methods typically deal with the generation based on a parameterized linear face model~\cite{BFM2009,Booth_2017_CVPR}, and now can generate high-fidelity faces in the virtual reality environment from single or multiple input photos~\cite{genova2018unsupervised,gecer2019ganfit}. However, the 3DMM and its variants are difficult to be applied to game environments. This is because most of the RPGs require effective interaction with their players, while the facial parameters generated by the 3DMM lack physical meaning, and the interactivity can not be guaranteed. Recently, Shi \etal~proposed Face-to-Parameter translation method (F2P)~\cite{Shi_2019_ICCV} to solve such problems and the generated faces can be easily applied to the game environments. In their method, they introduce a convolutional neural network to encode the face model so that the facial parameters can be easily optimized under a neural style transfer framework~\cite{Gatys2016Image}. On this basis, players can optionally manipulate the generated characters by manually adjusting their facial parameters. 

However, in the previous F2P methods~\cite{Shi_2019_ICCV,shi2020fast}, the expression parameters of the generated character and the facial identity parameters are highly coupled with each other, making it difficult to model the intrinsic facial features of the character. In the daily gaming experience, the players are more likely to accept the generated characters with neutral faces (expressionless characters). So far, the disentanglement of facial identity and expression in the character auto-creation still remains a challenge. For example, for players with very small eyes and players with their eyes are slightly closed, F2P based methods fail to distinguish the expression parameters of the two cases from their input photo effectively. Besides, the neural network based renderer in these methods is only designed for rendering front view faces, which makes these methods hard to be extended to multi-view rendering cases.

In this paper, considering the above problems, we propose a novel method for neutral face game character auto-creation. We formulate the neutral face generation under a multi-task self-supervised learning paradigm. Our method can be trained by maximizing the facial similarity between the rendered character and the input facial photo. To generate expressionless characters (``poker face'' characters), we deal with the disentanglement of the facial identity and expression with the help of a Generative Adversarial Network (GAN). We thus name our method as ``PokerFace-GAN''. To effectively measure the cross-domain similarity between the two faces, we design a multi-task loss function that considers multiple factors, including facial content, identity, pose, and facial landmarks. As suggested by Shi \etal, we further introduce an attention-based neural network named ``parameter translator'' to translate the above face image representations into three groups of parameters, i.e., ``expression parameters'', ``facial identity parameters'', and ``pose parameters''. These parameters either can be used for rendering 3D characters in the game environments or can be further manually fine-tuned by players. Considering that given a single input photo, the separation of the ``expression parameters'' and the ``facial identity parameters'' may have an infinite number of possible solutions, we introduce a discriminator to tackle this separation ambiguity problem. The discriminator is trained to tell whether the input facial parameters are from an expressionless face, in other words, whether the predicted facial identity parameters are coupled with expressions. Fig.~\ref{fig:cover} shows an example of the parameter separation of our method.\footnote{All face images shown in this paper are synthetic ones that are generated by using StyleGAN2~\cite{Karras2019stylegan2} to avoid violating ACM's copyright regulations.}

In addition to the adversarial training, we also take advantage of the differentiable rendering technique to make the rendering process of the game engine differentiable so that our method can be easily trained in an end-to-end fashion. Different from the previous methods that are built based on the neural renderer~\cite{Shi_2019_ICCV,shi2020fast}, we build a hard-programming based differentiable renderer which is more flexible to be applied for multi-view rendering. Our renderer integrates the advantages of the rendering process in both of the F2P and 3DMM methods, where we first build a 3D facial model with a large set of physically meaningful parameters, and then approximate the gradient of the rendering process based on triangle barycentric interpolation so that to ensure the gradients can be back-propagated smoothly from the rendered 2D images to the input end. Based on this improvement, our renderer can easily deal with expression basis and 3D-poses just like what the main-stream game engines can do. As a comparison, the neural renderers used in previous methods~\cite{Shi_2019_ICCV,shi2020fast} do not have such scalability.

Our contributions are summarized as follows:

1. We propose ``PokerFace-GAN'', a new method to automatically generate game characters with neutral faces (expressionless faces) based on a single input photo from players. We reformulate the character auto-creation problem as a disentangle problem of facial parameters and expression parameters. The problem is effectively solved with the help of adversarial training. To our best knowledge, there are few works have been done on this topic.

2. We introduce a hard-programming based differentiable character renderer, which not only preserves the full interactions required by games but also can be easily integrated into the training pipeline of our method. Our renderer can better deal with the multi-pose rendering problem than the previous differentiable rendering methods~\cite{Shi_2019_ICCV,shi2020fast}.

\begin{figure*}
  \includegraphics[width=0.9\linewidth]{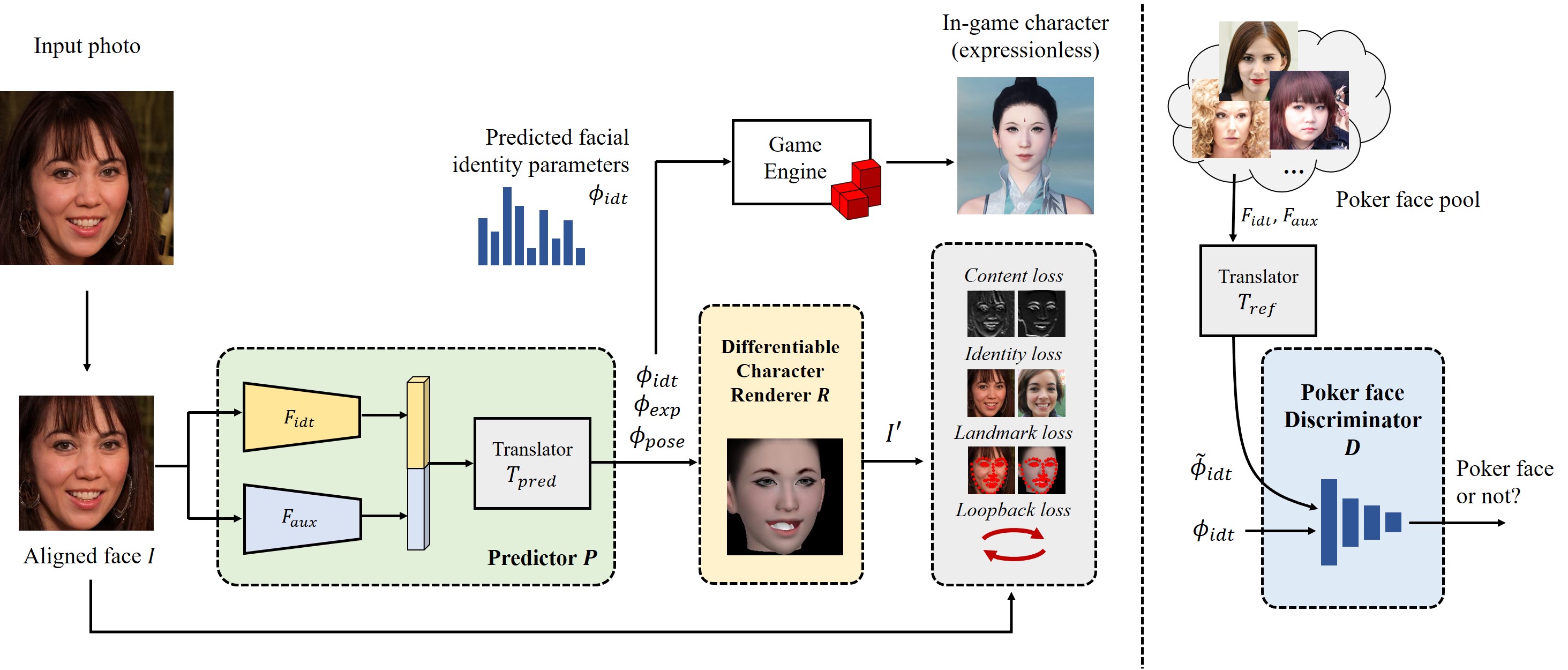}
  \caption{An overview of the proposed method. Our method contains three modules: 1) a Predictor $P$ which takes in an aligned facial photo $I$ and predict three groups of facial parameters (identity $\phi_{idt}$, expression $\phi_{exp}$, and pose $\phi_{pose}$); 2) a differentiable character renderer $R$ which imitates the behavior of the game engine to convert the predicted parameters to face images; and 3) a discriminator $D$ which is trained to classify whether the predicted facial parameters $\phi_{idt}$ contains expression or not.
  We also introduce multiple loss functions to measure the facial similarity between the rendered face and the real one.}
  \Description{framework}
  \label{fig:framework}
\end{figure*}

\section{Related Work}

{\bf 3D Face Reconstruction}. 3D face reconstruction, which focuses on building 3D Faces based on single or multiple 2D photos, has long been a fundamental but challenging task in computer vision. A representative of early monocular 3D face reconstruction is the 3DMM, which was originally proposed by Blanz \etal~in 1999~\cite{blanz1999morphable}. In this method, the authors first built a linear parameterized 3D Face model based on 3D scans and then solved the fitting problem between the parameterized 3D face and input photo~\cite{blanz1999morphable}. Recently, deep neural networks have greatly promoted the research progress of this field. Many methods adopt Convolutional Neural Networks (CNNs) to directly predict the parameters of 3DMM and are trained by minimizing the loss between the prediction and the pre-scanned faces~\cite{Tran_2017_CVPR,Dou_2017_CVPR,jackson2017large}. Thanks to the development of the self-supervised learning and differentiable rendering~\cite{kato2018neural} in the recent two years, neural networks are now able to generate 3D faces even without collecting the 3D scans ~\cite{Tewari_2017_ICCV,genova2018unsupervised}. In these methods, the facial identities, landmarks, and some other image features are introduced to measure facial similarity~\cite{tewari2017mofa,genova2018unsupervised}. Besides, to generate high-fidelity 3D faces, Generative Adversarial Networks (GANs) are also introduced to this task recently~\cite{gecer2019ganfit,lin2020towards}. Despite the recent advances in this field, most of the 3d face reconstruction methods are still difficult to be applied in RPGs since the morphable face model used in these methods is not friendly for user interactions.

{\bf Character Auto-Creation}. Character Auto-Creation is an emerging research topic recently that helps players to rapidly build their characters in RPGs. Considering that the characters in modern RPGs are designed to be precisely manipulated, Shi \etal~proposed a method called ``Face-to-Parameter translation'' (F2P) to predict the facial parameters of in-game characters~\cite{Shi_2019_ICCV}. In their method, the authors first train a generative network to make the game engine differentiable and then frame the auto-creation problem as a neural style transfer problem. Later, a fast and robust version of this method was proposed based on self-supervised training~\cite{shi2020fast}. Besides, the GANs also have been applied to solve the Character Auto-Creation problem. For example, Wolf \etal~proposed a method called ``Tied Output Synthesis (TOS)'' for building parameterized avatars based on an adversarial training framework~\cite{wolf2017unsupervised}. In this method, the authors first train a generator to select a set of pre-defined facial templates based on the input facial photo and then synthesize the avatar based on the selected templates. A common drawback of the above methods is that, these methods ignore the disentanglement of the expression parameters and the facial identity parameters.

{\bf Differentiable Rendering}. Rendering is a fundamental operation in computer graphics that converts 3D models into 2D images. Traditional rendering methods used in 3D graphics only consider the forward process and are non-differentiable due to a discrete processing step called ``rasterization''. Recently, differentiable rendering has drawn great attention and is becoming a research hot-spot in computer graphics. One of the earliest differentiable renderers was proposed by Loper \etal, named OpenDR~\cite{loper2014opendr}, where the authors use the first-order Taylor expansion to approximate the gradient of the rasterization. Some recent improvements of the differentiable rendering include the interpolation-based methods~\cite{kato2018neural,Liu_2019_ICCV}, triangle barycentric based method~\cite{genova2018unsupervised}, and the Monte Carlo based method~\cite{li2018differentiable}. Besides, since deep neural networks are naturally differentiable, a new research topic called ``neural rendering'' was quickly introduced to a variety of rendering tasks, such as 3D object reconstruction~\cite{yan2016perspective,nguyen2018rendernet,eslami2018neural}, 3D avatar generation~\cite{wolf2017unsupervised,Shi_2019_ICCV,shi2020fast}, etc.

\section{Methodology}

Fig.~\ref{fig:framework} shows an overview of the proposed method. Our method consists of three modules: a predictor $P$ for facial parameter prediction, a differentiable character renderer $R$ for converting facial parameters to 3D characters, and a discriminator $D$ for classifying whether the predicted facial parameters contain expressions. We also design multiple loss functions to measure the facial similarity between the generated character and the input facial photo. The measurement is conducted under a self-supervised learning framework where we enforce the facial representation of the rendered image $I^\prime$ to be similar to that of its input facial photo $I$: $I^\prime = R(P(I)) \rightarrow I$.

\begin{figure}
  \includegraphics[width=0.9\linewidth]{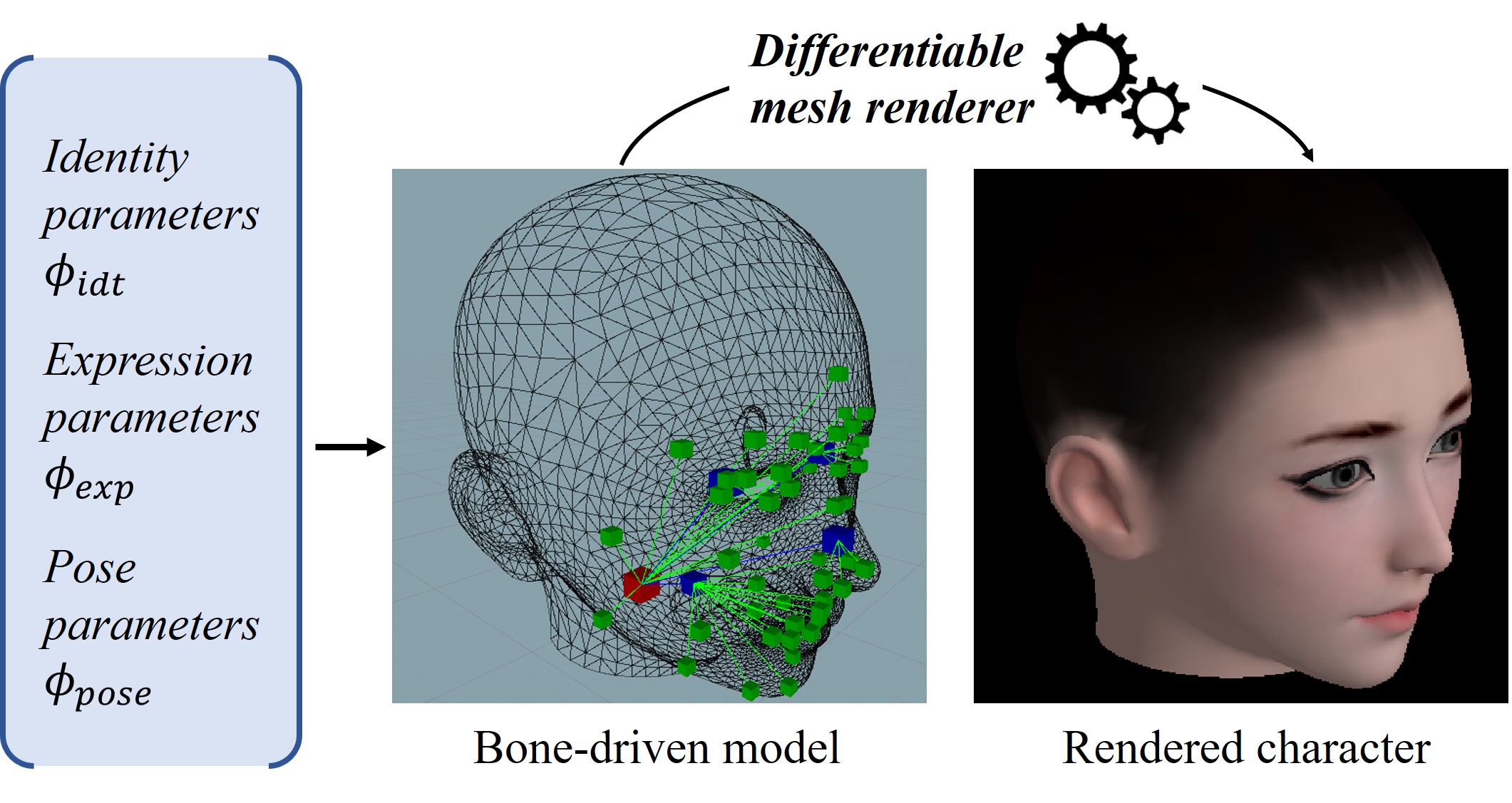}
  \caption{The rendering pipeline of our differentiable character render. As what most game engines do, we first transform the facial parameters (identity $\phi_{idt}$, expression $\phi_{exp}$, and pose $\phi_{pose}$), to a bone-driven face model. Then we convert the model to 3D meshes and finally produce the rendering result based on the mesh renderer~\cite{genova2018unsupervised}.}
  \Description{differentiable character creator}
  \label{fig:dcc}
\end{figure}

\subsection{Differentiable Character Renderer}

We build our differentiable character renderer $R$ to simulate the rendering behavior of the game engine. As shown in Fig.~\ref{fig:dcc}, our renderer takes in three groups of facial parameters - 1) the facial identity parameters $\phi_{idt}$, 2) expression parameters $\phi_{exp}$, and 3) pose parameters $\phi_{pose}$, and then converts these parameters to a 2D face image $I^\prime$:
\begin{equation}
I^\prime = R(\phi_{idt}, \phi_{exp}, \phi_{pose}).
\label{eq:dcc}
\end{equation}
The rendering process mainly includes the following three steps:

1. We first transform the \emph{facial identity parameters} $\phi_{idt}$ to a set of bone parameters based on the linear relations defined by game developers. In the bone-driven model, bones are in tree-like structures and each bone is attached with 9 parameters representing local translation, rotation, and scale, respectively.

2. We then compute the coordinates of the mesh vertices in the world system by using the local transformation matrix (determined by the bone parameters) hierarchically from the current bone to its root. Besides, we also integrate the expression parameters $\phi_{exp}$ at this step where the expression basis is added to the 3D mesh before the coordinate transformation.

3. After obtaining the 3D mesh of a character, we attach the pose parameters $\phi_{pose}$ and make transformation on 3D mesh. We finally use the mesh renderer~\cite{genova2018unsupervised} to produce the character. We choose this renderer because human faces typically have smooth surfaces and this renderer focuses more on the inner part of 3D mesh rather than other differentiable renderers.

For simplicity, the hair-style and the background are ignored during the rendering process. For more computation details of coordinate transformation, please refer to our supplementary materials.

\begin{figure}
  \includegraphics[width=0.9\linewidth]{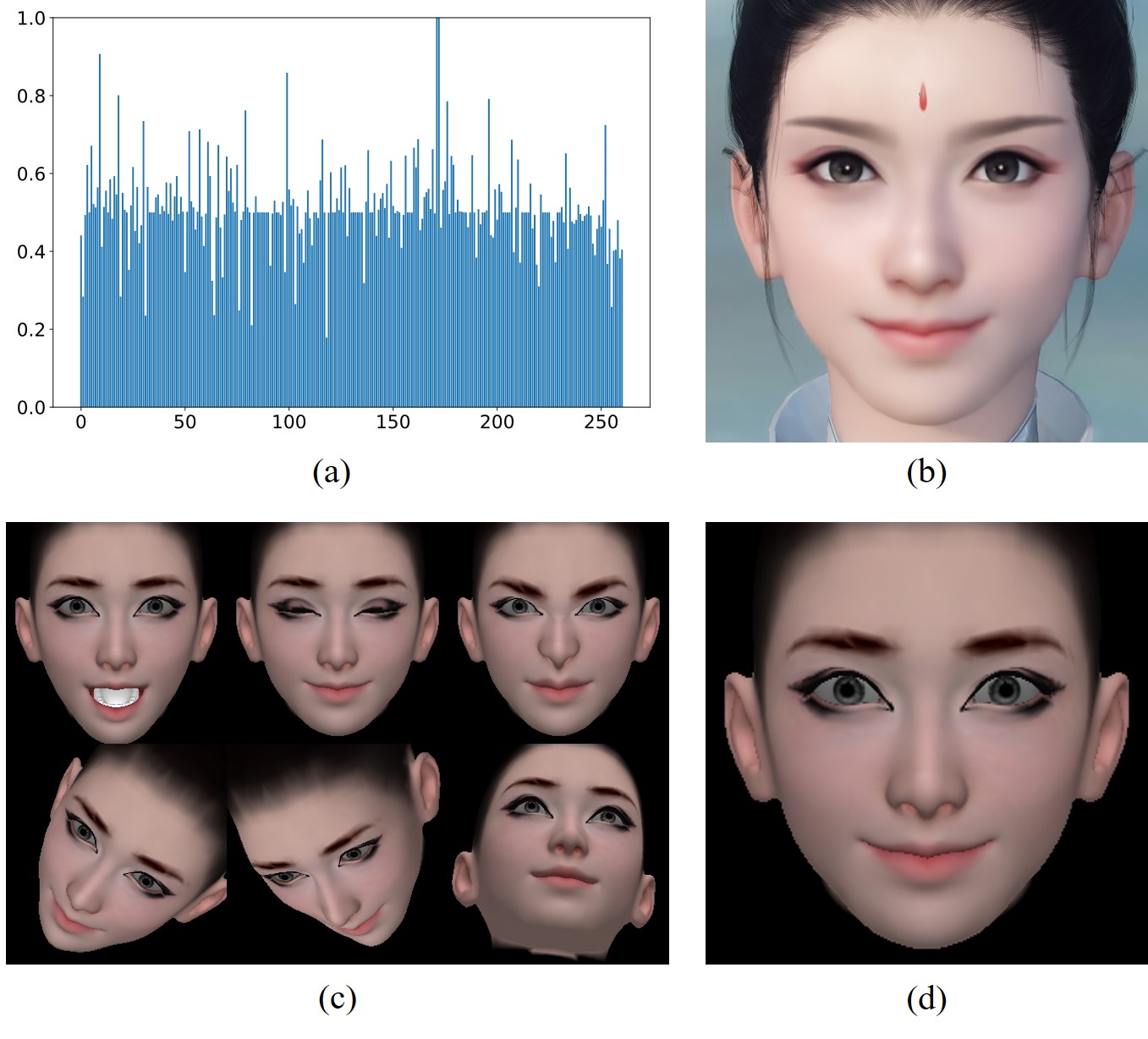}
  \caption{A comparison between the renderer proposed in~\cite{Shi_2019_ICCV} and the one proposed in this paper. Given the same set of facial parameters, the renderer in~\cite{Shi_2019_ICCV} can only generate a front-view face image (b), while the proposed one can generate faces with different poses and expressions (c-d).}
  \Description{RendererCompare}
  \label{fig:render}
\end{figure}

\subsection{Predictor}

Our predictor $P$ consists of three sub-networks: two facial feature extractors $F_{recg}$ and $F_{aux}$, and a parameter translator $T_{pred}$.

{\bf Facial feature extractor} $F_{recg}$: A key to learning effective face similarity measurement is to extract an identity-variant facial description for players. We thus introduce a well-known face recognition network named LightCNN-29v2~\cite{wu2018light} to extract a group of 256-d facial identity features for each input face image. 

{\bf Facial feature extractor} $F_{aux}$: To generate more precise facial parameters like expressions and poses, we further introduce an auxiliary network to compensate for the ignored information by the LightCNN-29v2 and enrich the facial features. We use the ResNet-50~\cite{he2016deep} as the backbone of the $F_{aux}$ and generate another group of 256-d auxiliary embeddings. The feature representations from the $F_{recg}$ and $F_{aux}$ are finally concatenated together and fed to the parameter translator $T_{pred}$ for generating facial parameters.

{\bf Parameter translator} $T_{pred}$: We adopt the attention-based multi-layer perceptron~\cite{genova2018unsupervised,shi2020fast} as our parameter translator to convert above embeddings to the corresponding facial parameters. We modified the prediction head for each group of facial embeddings, and train our translator $T$ to predict the three groups of facial parameters ( $\phi_{idt}$, $\phi_{exp}$, and $\phi_{pose}$) separately. The translation process can be represented as follows:
\begin{equation}
\begin{split}
    \phi_{idt}, \phi_{exp}, \phi_{pose} &= P(I) \\
    &=T_{pred}([F_{recg}(I);F_{aux}(I)]),
\end{split}
\end{equation}
where the $[ \ ; \ ]$ represents the feature concatenation operation.

\subsection{PokerFace-GAN}

\begin{figure}[t]
  \includegraphics[width=0.9\linewidth]{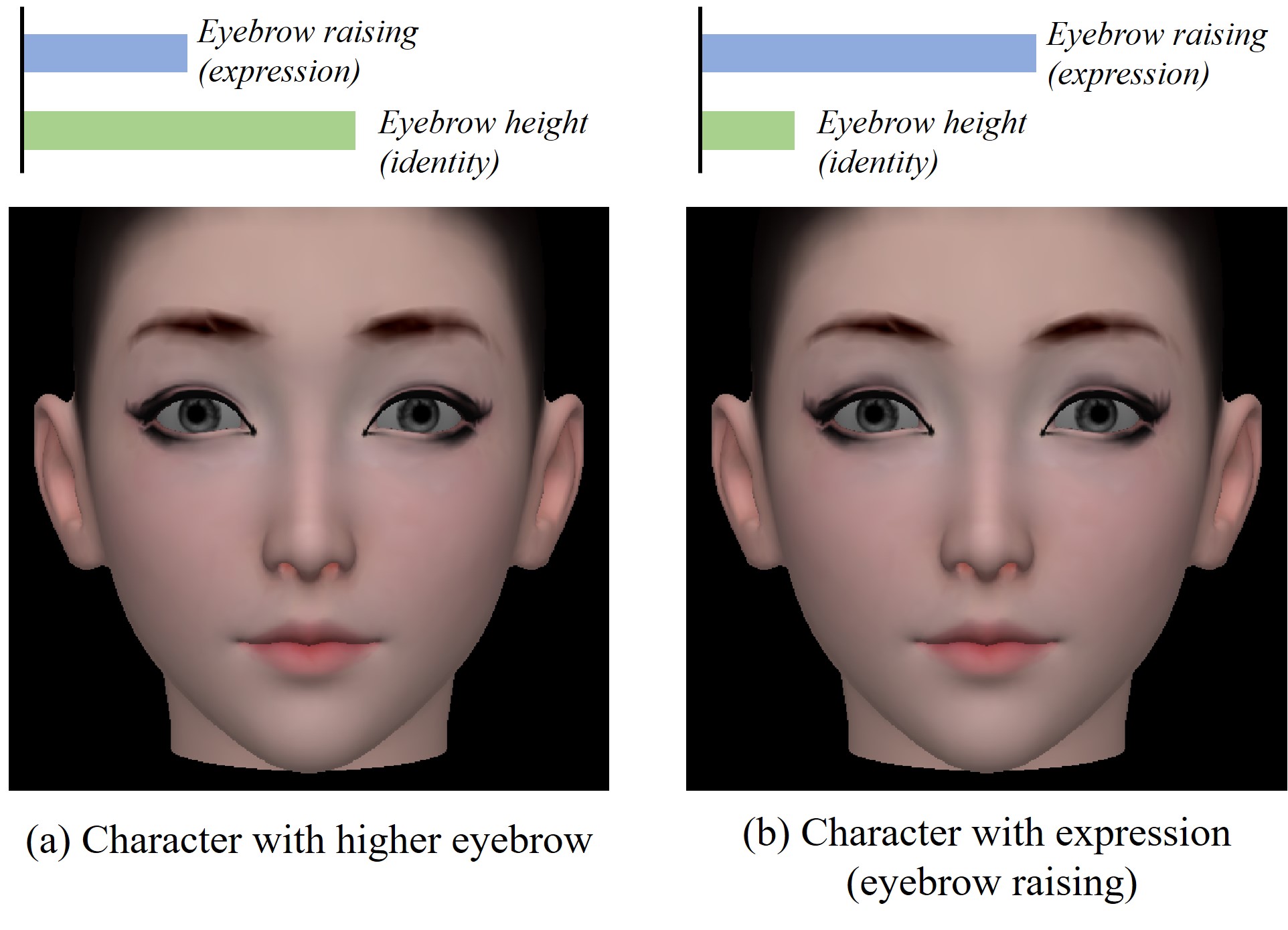}
  \caption{A simple illustration of the parameter coupling problem in the character auto-creation task - although the characters in (a) and (b) look similar, they are rendered from very different parameters.}
  \Description{a coupling example}
  \label{fig:mix}
\end{figure}

Different from the 3DMM~\cite{blanz1999morphable} and its variants~\cite{Tran_2017_CVPR,Dou_2017_CVPR,jackson2017large} where the facial parameters were decoupled from the beginning, in many facial customization systems, the facial identity and expressions are highly coupled with each other. For example, as shown in Fig.~\ref{fig:mix}, to raise the eyebrow of an in-game character, we can either edit the identity parameters to change the height of the eyebrow (shown in Fig.~\ref{fig:mix} (a)), or edit the expression parameters (shown in Fig.~\ref{fig:mix} (b)). As a result, although the two faces are rendered from different parameters, they look very similar.

In our task, the disentanglement of the facial identity and the expression can be essentially viewed as a mixed source separation problem. Recently, adversarial training shows its power on these problems~\cite{zoudeep}. We thus introduce adversarial training to our framework, where a discriminator $D$ is trained to classify whether the generated facial identity parameters contain any expressions or not. We then feed the output of the discriminator back to the predictor $P$ to further improve it and make the generated parameters indistinguishable.

Specifically, we firstly build a ``poker face pool'', where the images in this pool are all from front-view, expression-less faces. We then build another reference translator $T_{ref}$ which shares the same architecture as the $T_{pred}$. We train the $T_{ref}$ to extract expressionless parameters from this pool: 
\begin{equation}
    \Tilde{\phi}_{idt}, \Tilde{\phi}_{pose}=T_{ref}([F_{recg}(I);F_{aux}(I)]).
\end{equation}
Note that the $T_{pred}$ produces three groups of parameters (identity, expression, and pose), while the reference translator $T_{ref}$ is trained to predict the identity and pose only. This design ensures that the reference parameters computed from the poker face pool contain absolutely zero expressions. 

We build a four-layer multi-layer perceptron as our discriminator. The discriminator takes the facial identity parameters $\Tilde{\phi}_{idt}$ produced by the reference translator $T_{ref}$ as positive samples, and takes those $\phi_{idt}$ from the $T_{pred}$ as negative ones. The objective function of the adversarial training can be written as follows:
\begin{equation}
\begin{split}
    \mathcal{L}_{gan}(P,D) &= \mathbb{E}_{I \sim p_{pool}(I)} {\log D(T_{ref}(f(I)))} \\
     &+ \mathbb{E}_{I \sim p_{data}(I)} \{\log(1 - D(T_{pred}(f(I))))\},
\end{split}
\label{eq:gan}
\end{equation}
where $f(I) = [F_{recg}(I);F_{aux}(I)]$ represents the features extracted by the facial feature extractors $F_{recg}$ and $F_{aux}$. The training of the PokerFace-GAN is a minimax optimization process, where the $P$ tries to minimize this objective while $D$ tries to maximize it: $P^\star = \arg \min_P \max_D \mathcal{L}_{gan}(P,D)$.

\subsection{Facial Similarity Measurement}

Based on the differentiable character renderer $R$ and the predictor $P$, the character auto-creation can be viewed as a facial similarity measurement problem, where the predictor is trained to maximize the facial similarity between the rendered face $I^\prime$ and an input real one $I$. 

Considering the domain gap between the game character and the real facial photo, in addition to the facial identity loss that is commonly used in 3D face reconstruction, we further integrate facial content loss~\cite{Shi_2019_ICCV}, landmark loss, and loop-back loss~\cite{genova2018unsupervised} to our framework to improve the facial similarity measurement and generalization ability.

{\bf Facial identity loss}.
We adopt the same face recognition network LightCNN-29v2 used in our predictor $P$ to compute the identity variant face representations, where a 256-d facial embedding is generated. We use cosine distance of two facial embeddings to measure the similarity of two face images, and the identity loss can be defined as follows:
\begin{equation}
\mathcal{L}_{idt}(I, I^\prime) = 1 - cos<F_{recg}(I), F_{recg}(I^\prime)>.
\label{eq:loss_idt}
\end{equation}

{\bf Facial content loss}.
The facial content loss provides constraints on the contour and displacement of different face components in the two images. Due to the style gap between the in-game character and the real facial photo, instead of computing the similarity based on the raw pixel space, we follow Shi \etal~and define the similarity based on the facial segmentation outputs. We first train a facial segmentation network to extract style-irrelevant local facial representation. Then we define the content loss as the $l_1$ distance between the middle-level features of two faces:
\begin{equation}
\mathcal{L}_{ctt}(I, I^\prime) = {||F_{seg}(I) - F_{seg}(I^\prime)||}_1.
\label{eq:loss_ctt}
\end{equation}
The facial segmentation network is built based on the Resnet-50 backbone. We train this network on the Helen dataset~\cite{le2012interactive}. To produce pixel-wise output, the fully connected layers in the Resnet-50 are replaced by a $1\times 1$ convolutional layer. The convolutional strides in the last two stages are also removed to increase the resolution of the segmentation result.

{\bf Landmark loss}.
Considering that facial content loss are pose-sensitive, here we further design a landmark loss to improve the robustness of the loss on pose changes. We first use the dlib library~\cite{dlib09} to extract 68 facial landmarks for each input face image. Then, given a group of predicted facial parameters, we mark the same number of vertices according to the dlib landmarks and project them into a 2D plane when rendering. We define the landmark loss by using the weighted $l_1$ distance between the landmarks of the input photo and the rendered game character: 
\begin{equation}
\mathcal{L}_{lm}(I, I^\prime) = {||\omega P_{lm} - \omega P_{lm}^\prime||}_1,
\label{eq:loss_lm}
\end{equation}
where $P_{lm}$ is the landmarks extracted by dlib on input photo, $P_{Lm}^\prime$ is the landmarks extracted from the game character. $\omega$ is the point-wise weight. We put more weights on the landmarks located on the eyes, nose, and mouth to encourage a more precise correspondence between these regions in the two faces.

{\bf Loop-back loss}. 
We follow Genova \etal~\cite{genova2018unsupervised} and define the loop-back loss to ensure our predictor $P$ correctly interprets its own output. After we obtain the rendered image $I^\prime$, we further feed it into our Predictor $P$ to produce three groups of new facial parameters $\phi_{idt}^\prime, \phi_{exp}^\prime, \phi_{pose}^\prime=P(I^\prime)$ and then force the generated parameters before and after the loop unchanged. We use $l_1$ distance to measure the loss between the two groups of parameters. Our loop-back loss can be written as follows:
\begin{equation}
    \mathcal{L}_{loop}(I, I^\prime) = {||[\phi_{idt}, \phi_{exp}, \phi_{pose}] - [\phi_{idt}^\prime, \phi_{exp}^\prime, \phi_{pose}^\prime]||}_1.
\label{eq:loss_loop}
\end{equation}

{\bf Final loss function}. Our final loss function is defined by combining the GAN loss (\ref{eq:gan}) and the facial similarity losses (\ref{eq:loss_idt}), (\ref{eq:loss_ctt}), (\ref{eq:loss_lm}), and (\ref{eq:loss_loop}). Since we have two translators $T_{pred}$ and $T_{ref}$, we apply the facial similarity loss functions to each of them. The combined loss function can be written as follows:
\begin{equation}
\begin{split}
\mathcal{L}_s(P,D) &= \lambda_{idt} (\mathcal{L}_{idt}(I,I_p^\prime) + \mathcal{L}_{idt}(I,I_r^\prime))\\
        & + \lambda_{ctt} (\mathcal{L}_{ctt}(I,I_p^\prime) + \mathcal{L}_{ctt}(I,I_r^\prime))\\
        & + \lambda_{lm} (\mathcal{L}_{lm}(I,I_p^\prime) + \mathcal{L}_{lm}(I,I_r^\prime))\\
        & + \lambda_{loop} (\mathcal{L}_{loop}(I,I_p^\prime) + \mathcal{L}_{loop}(I,I_r^\prime))\\
        & + \lambda_{gan} \mathcal{L}_{gan}
\end{split}
\label{eq:loss_similarity}
\end{equation}
where $I_p^\prime$ and $I_r^\prime$ are the images rendered based on the parameters predicted by the $T_{pred}$ and $T_{ref}$, respectively.

We further add multiple regularization terms on the above losses: 
\begin{equation}
\begin{split}
\mathcal{L}_r(P) & = \beta_{idt}({||\phi_{idt}||}^2_2 + {||\Tilde{\phi}_{idt}||}^2_2) + \beta_{exp}{||\phi_{exp}||}^2_2\\
    &+ \beta_{pose}({||\phi_{pose}||}^2_2 + {||\Tilde{\phi}_{pose}||}^2_2) \\
    & + \beta_{image}{||I_p^\prime - I_r^\prime||}_1 + \beta_{ref}{||\phi_{idt} - \Tilde{\phi}_{idt}||}_1.
\end{split}
\label{eq:loss_regular}
\end{equation}

Finally, we aim to solve the following mini-max optimization problem:
\begin{equation}
\begin{split}
P^\star, D^\star = \arg\min_P\max_D \mathcal{L}_s(P, D) + \mathcal{L}_r(P).
\end{split}
\label{eq:minmax}
\end{equation}

\subsection{Implementation Details}

A complete implementation pipeline of our method is summarized as follows:
\begin{itemize}
\item {\bf Training Phase (I).} Train the face recognition network $F_{recg}$ and the face segmentation network $F_{seg}$.
\item {\bf Training Phase (II).} Freeze $F_{recg}$, $F_{seg}$, and train the predictor $P$ and the discriminator $D$ according to Eq.~(\ref{eq:minmax}).
\item {\bf Inference Phase.} Given an input photo $I$, predict the facial parameters: $\phi_{idt}, \phi_{exp}, \phi_{pose}=P(I)$.
\end{itemize}

We train our predictor $P$ and discriminator $D$ on the CelebA dataset~\cite{liu2015faceattributes} with the Adam optimizer~\cite{kingma2014adam}. We set the learning rate = $10^{-4}$ and max-iteration = 10 epochs. We select a small subset of images with front-view poker faces ($\sim$ 5,000 photos) to build our poker face pool. During the training phase II, we remove the loss of the discriminator $D$ in the first epoch to initialize the neutral face pool with the reference translator. In the rest epochs, we enable the discriminator loss and continuously update this pool. When training the face recognition network $F_{recg}$ and the facial segmentation network $F_{seg}$, we follow the training configuration of Shi \etal~\cite{shi2020fast}.

In the landmark loss, the weights $\omega$ on eyes, nose, and mouth are set to 5.0, and the rests are set to 1.0. In our combined loss function, we set $\lambda_{idt} = 0.05$, $\lambda_{ctt} = 1$, $\lambda_{lm} = 2$, and $\lambda_{loop} = 2$. In our regularization terms, we set $\beta_{idt} = 1$, $\beta_{exp} = 0.1$, $\beta_{pose} = 0.1$, $\beta_{image} = 0.1$, and $\beta_{ref} = 0.1$.

The dimensions of the facial parameters $\phi_{idt}$, $\phi_{exp}$, and the $\phi_{pose}$ are set to 261, 22, and 6, respectively. Among these parameters, $\phi_{idt}$ defines the translation, rotation, and scaling factors of 29 facial parts. $\phi_{exp}$ defines the intensity of 22 expression basis. $\phi_{pose}$ defines the translation and rotation of the face on x, y, and z axis. For more details on the facial parameters, please refer to our supplementary materials .

Our code is written based on PyTorch~\cite{PyTorch_NEURIPS2019_9015}. For the differentiable renderer, since the authors of the mesh-renderer~\cite{genova2018unsupervised} only released their TensorFlow version, we re-implement it in PyTorch, speed up the rendering with CUDA, and also make this version publicly available\footnote{https://github.com/FuxiCV/pt\_mesh\_renderer}.

\begin{figure}[t]
  \includegraphics[width=0.95\linewidth]{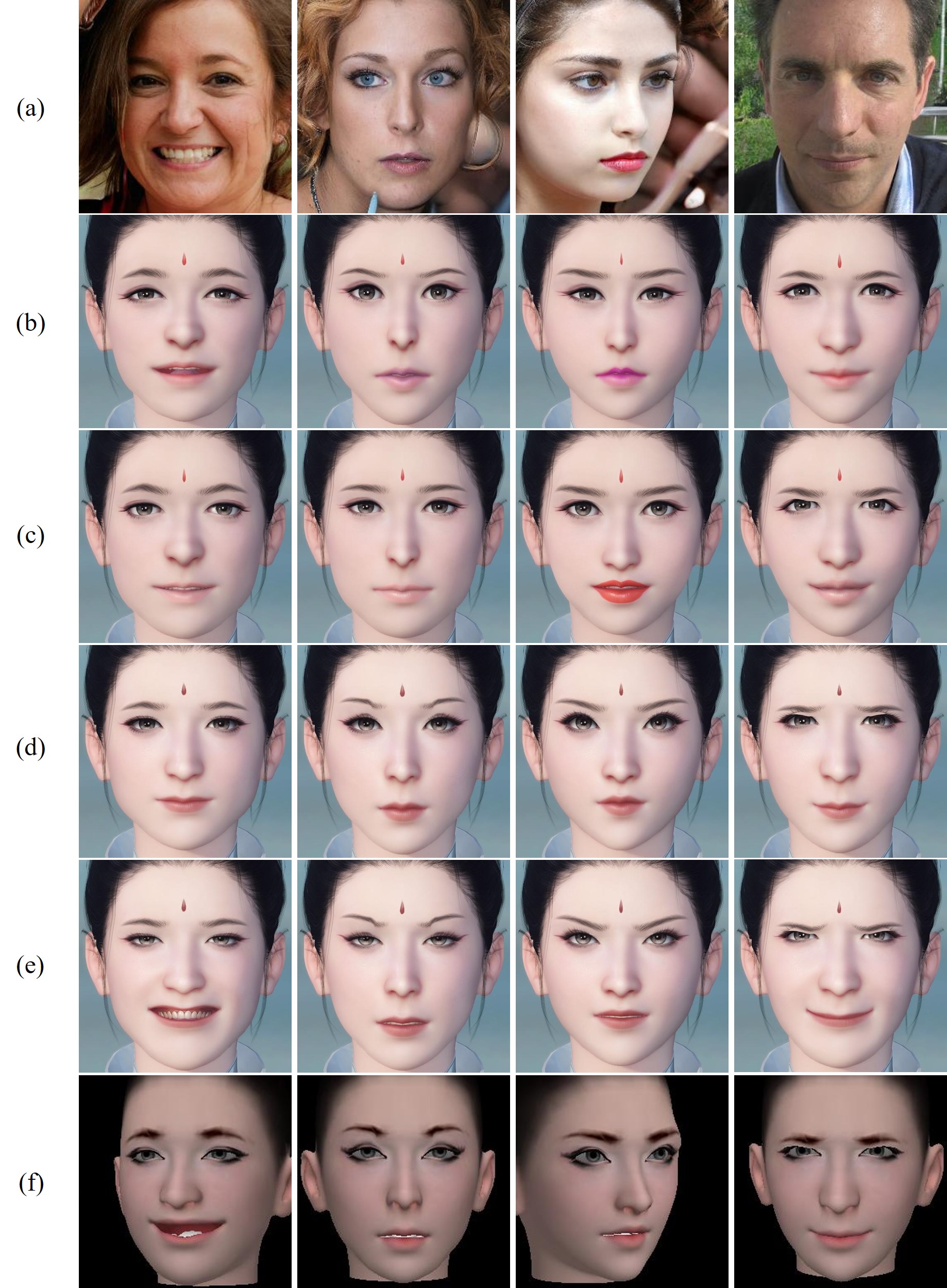}
  \caption{A visual comparison of different methods for in-game character auto-creation. (a) Aligned input photos, (b) The results of F2P~\cite{Shi_2019_ICCV}, (c) The results of FR-F2P~\cite{shi2020fast}, (d) Our results (identity-only) (e) Our results (with both identity and expression) (f) Our results rendered by our differentiable character renderer.}
  \Description{comparison}
  \label{fig:big}
\end{figure}

\begin{table*}[t]
\centering
\caption{A comparison of different methods for character creation. The accuracy is computed on seven face verification datasets: LFW \cite{LFW}, CFP\_FF \cite{CFP}, CFP\_FP \cite{CFP}, AgeDB \cite{Agedb}, CALFW \cite{CALFW}, CPLFW \cite{CPLFW}, and Vggface2\_FP \cite{vggface2}. We follow the evaluation metric of the benchmark ``face.evoLVe''~\cite{zhao2019look}. Higher scores indicate better (\emph{We follow the 3DMM-CNN and normalize the face embeddings by PCA for a fair comparison}).}
\begin{tabular}{l|ccccccc|c}
\toprule
\multirow{2}{*}{\textbf{Method}} & \multicolumn{7}{c|}{\textbf{Datasets}}& \multirow{2}{*}{\textbf{Speed$^*$}}\\ 
 & LFW & CFP\_FF & CFP\_FP & AgeDB & CALFW & CPLFW & Vggface2\_FP & \\
\midrule 
3DMM-CNN~\cite{Tran_2017_CVPR} & 0.9235 & - & - & - & - & - & - & $\sim10^2$Hz \\
F2P~\cite{Shi_2019_ICCV} & 0.6977 & 0.7060 & 0.5800 & 0.6013 & 0.6547 & 0.6042 & 0.6104 & $\sim1$Hz \\
FR-F2P~\cite{shi2020fast} & 0.9402 & 0.9450 & 0.8236 & 0.8408 & 0.8463 & 0.7652 & 0.8190 & $\sim10^3$Hz \\
Ours & {\bf 0.9550} & {\bf 0.9587} & {\bf 0.8457} & {\bf 0.8573} & {\bf 0.8750} & {\bf 0.7857} & {\bf 0.8382} & $\sim 600$Hz \\
\midrule
LightCNN-29v2~\cite{wu2018light} & 0.9958 & 0.9940 & 0.9494 & 0.9597 & 0.9433 & 0.8857 & 0.9374 & $\sim10^3$Hz\\
\bottomrule
\multicolumn{9}{l}{* Inference speed under GTX 1080Ti. The time cost for model loading and face alignment are not considered.}\\
\end{tabular}%
\label{tab:comparison}
\end{table*}%

\begin{table*}[t]
\centering
\caption{Ablations analysis on the proposed two technical components in our method: 1) the facial feature extractor $F_{aux}$, and 2) adversarial training. We observe a noticeable accuracy drop when we remove either of the two components.}
\begin{tabular}{l|ccccccc}
\toprule
 \multirow{2}{*}{\textbf{Ablations}} & \multicolumn{7}{c}{\textbf{Datasets}}\\
 & LFW & CFP\_FF & CFP\_FP & AgeDB & CALFW & CPLFW & Vggface2\_FP \\
\midrule
Ours (w/o $F_{aux}$) & 0.9285 & 0.9294 & 0.8081 & 0.8333 & 0.8360 & 0.7498 & 0.8042  \\
Ours (w/o adv. training) & 0.9360 & 0.9324 & 0.8080 & 0.8378 & 0.8392 & 0.7508 & 0.8044  \\
Ours (full implementation) & {\bf 0.9428} & {\bf 0.9480} & {\bf 0.8244} & {\bf 0.8462} & {\bf 0.8580} & {\bf 0.7648} & {\bf 0.8238} \\
\midrule 
LightCNN-29v2~\cite{wu2018light} & 0.9948 & 0.9939 & 0.9476 & 0.9537 & 0.9438 & 0.8872 & 0.9326 \\
\bottomrule
\end{tabular}%
\label{tab:ablation}
\end{table*}%

\section{Experiments}

We test our method on a role-playing game named ``Justice'' (\url{https://n.163.com}), which was launched in June 2018 and now has over 20 million registered users. The game is mainly developed for East Asian gamers. In our experiment, all training images are from the CelebA dataset~\cite{liu2015faceattributes} and visualized test cases are generated by StyleGAN2~\cite{Karras2019stylegan2}. 

\subsection{Disentanglement of identity and expression}

Fig.~\ref{fig:big} shows a group of generation results of our method. We can see that our method not only generates vivid game characters that look like the input photo (with both identity and expression) but can also generate expressionless characters (identity only). The players can thereby obtain a cleaner character. We can also see that the characters generated by previous methods can have rich expressions, which is usually not desired by players.

\begin{figure*}[ht]
  \includegraphics[width=0.9\linewidth]{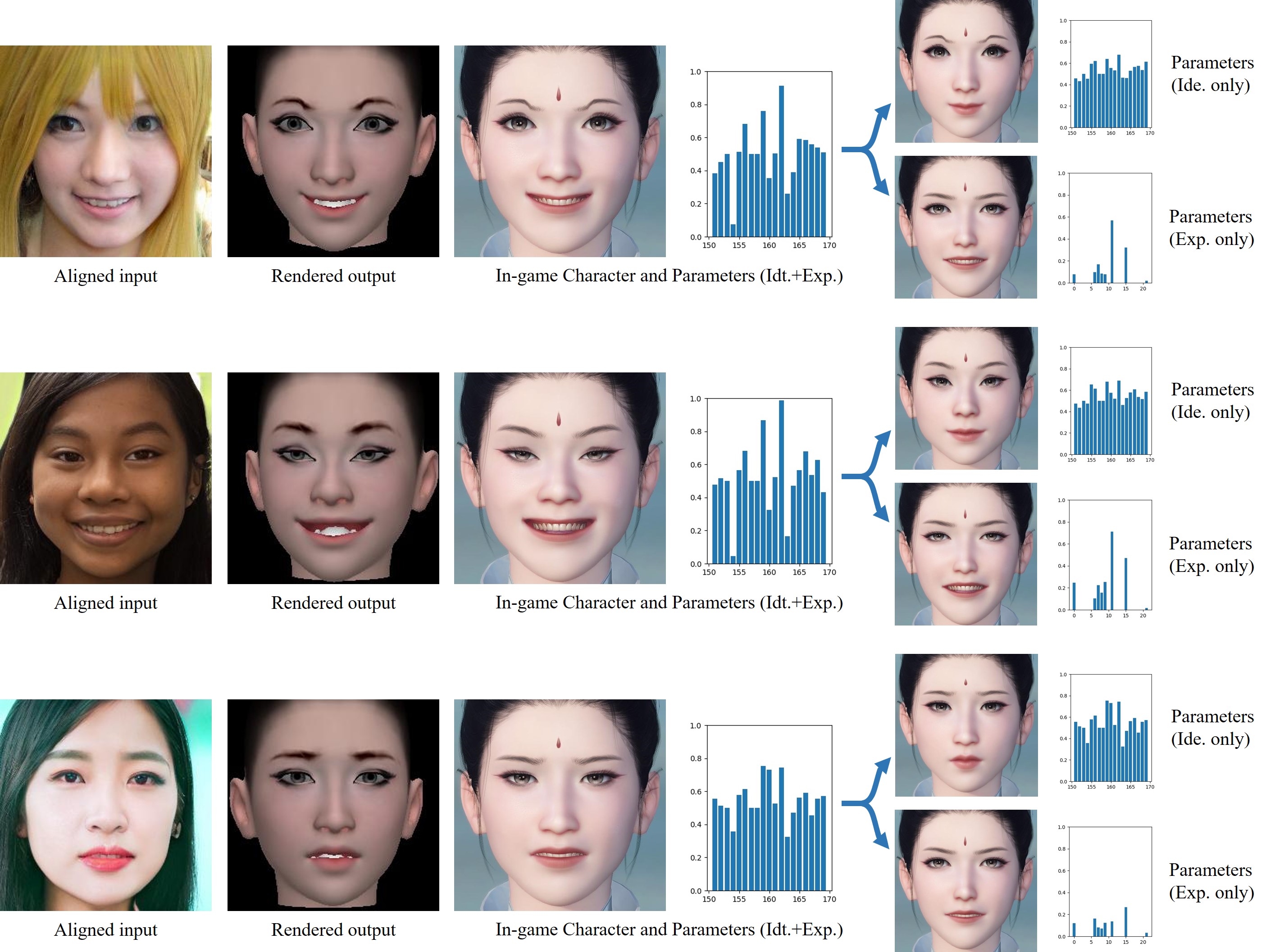}
  \caption{Some examples of the separation results of the identity (Idt.) and expression (Exp.) by using our method. Notice that there are in total of 261 facial identity parameters for each face, for better visualization, we only plot its 151 $\sim$ 170 dimensions (the mouth part).}
  \Description{details}
  \label{fig:details}
\end{figure*}

In Fig.~\ref{fig:details}, we show that by introducing PokerFace-GAN, our final prediction can disentangle the expression parameters from facial parameters. For example, in the last group of Fig.~\ref{fig:details}, the input photo contains a mouth-open, slightly eye-closed face. We can see there are significant changes between the facial parameters before and after disentanglement.

\subsection{Comparison with other methods}

We compare our method with some recent methods for game character auto-creation, include the F2P~\cite{Shi_2019_ICCV}, FR-F2P~\cite{shi2020fast}, and 3DMM-CNN~\cite{Tran_2017_CVPR}, on both of their accuracy and speed.

To make quantitative comparisons, we test our method on seven large face verification datasets, including the LFW \cite{LFW}, CFP\_FF \cite{CFP}, CFP\_FP \cite{CFP}, AgeDB \cite{Agedb}, CALFW \cite{CALFW}, CPLFW \cite{CPLFW}, and Vggface2\_FP \cite{vggface2}. We follow the previous works~\cite{Tran_2017_CVPR,Shi_2019_ICCV,shi2020fast} and use the accuracy on the face verification task as our evaluation metric. The reason why the face verification can reflect the quality of character auto-creation is that if the two faces are from the same person, they should have similar facial parameters. For doing this, we first generate facial identity parameters for every pair of input faces in each face verification dataset. Then we use the ``face.evoLVe''~\cite{zhao2019look,zhao2019multi,zhao2018towards}, a face verification benchmark toolkit, to compute the accuracy based on the facial (identity) parameters. Table~\ref{tab:comparison} shows the evaluation results on the seven datasets, where a higher score indicates a better performance. The results of other methods are reported by previous literature~\cite{Tran_2017_CVPR,Shi_2019_ICCV,shi2020fast}. Note that although the 3DMM-CNN~\cite{Tran_2017_CVPR} cannot generate physical meaningful facial parameters, here we still list its result as a reference. We also list the performance of LightCNN-29v2 on input photos as a reference (upper-bound accuracy), because it is performed directly in its identity representations, while other scores are computed based on the generated facial parameters of 3D characters. It can be seen that our method achieves the highest score among all datasets. Besides, the processing speed of our method is much faster than the 3DMM-CNN~\cite{Tran_2017_CVPR} and F2P~\cite{Shi_2019_ICCV}, and is comparable to the FR-F2P~\cite{shi2020fast}.


\subsection{Ablation study}

In this experiment, we analyze the importance of each technical component of the proposed method and how they contribute to the result. Considering that in the previous literature~\cite{genova2018unsupervised,shi2020fast}, the authors have verified the effectiveness of the parameter translator and the facial similarity loss functions, here we only focus on verifying the importance of our newly proposed components, including:

1) the facial feature extractor $F_{aux}$;

2) the discriminator $D$ (adversarial training);

We use the same quantitative evaluation metric as we used in our comparison experiment. Our full implementation is first evaluated. Then we gradually remove the above two components one by one. Table~\ref{tab:ablation} shows their accuracy on different datasets. We can see that if we remove either of the facial feature extractor $F_{aux}$ and the discriminator $D$ from our method, there will be a consistent drop in the face verification accuracy on all the seven datasets. Besides, the removal of the discriminator brings a more noticeable accuracy drop on our method. This suggests that adversarial training not only helps the disentanglement of the identity and expression but also improves the accuracy of the predicted facial parameters.

\begin{figure}[ht]
  \includegraphics[width=\linewidth]{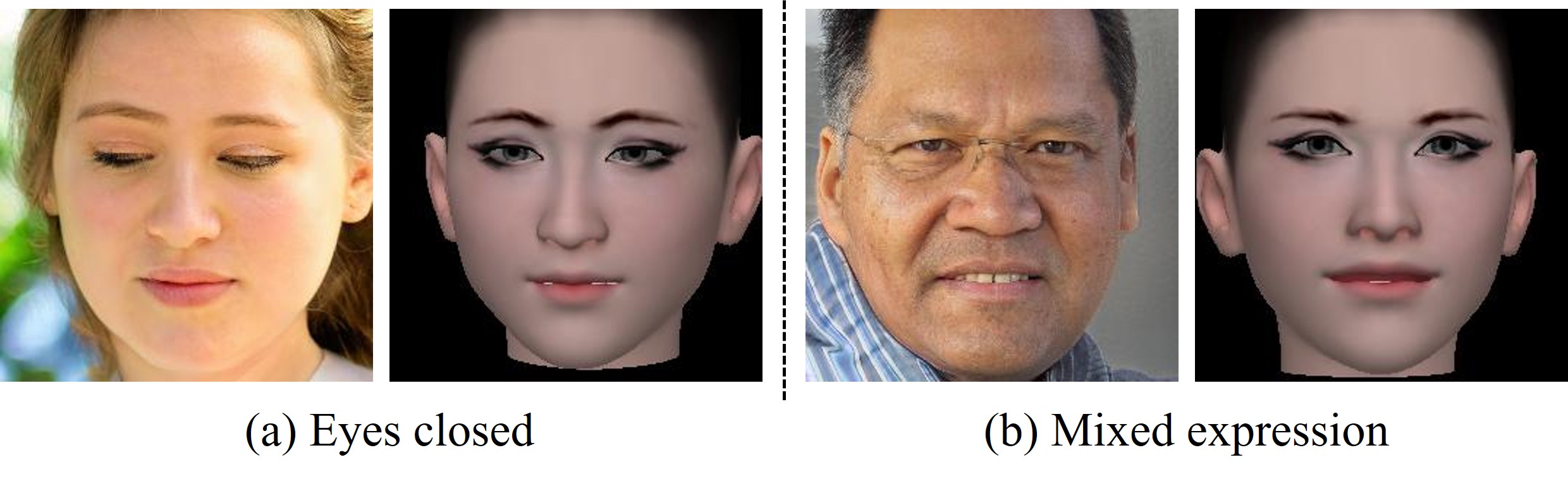}
  \caption{Failure cases of our method: input faces with (a) eyes closed, and (b) mixed expressions.}
  \Description{badcase}
  \label{fig:badcase}
\end{figure}

\subsection{Failure cases}

Fig.~\ref{fig:badcase} shows two failure cases of our method. The first one is that our method may fail when the eyes of the input facial photo are closed. This may be because the landmarks extracted by dlib are not accurate in this condition. The second one is that it is hard for our method to deal with the faces with complex or mixed expressions. This is because most of the face samples in our training set (CelebA) have no expression or laugh only. This problem can be alleviated by training on larger and more complex face datasets. Since the training of our method is self-supervised and does not require any annotations on the faces, our method can be easily extended to other datasets.

\section{Conclusion}

In this paper, we propose a new method for the auto-creation of in-game characters. Our method takes advantage of the adversarial training to effectively disentangle the expression from the identity and thus can generate characters with neutral (expressionless) faces. We also build a hard-programming based differentiable character rendering to make the auto-creation process differentiable, which is more flexible and more effective in rendering multi-view faces than the previous renderers. Our method achieves better generation accuracy compared to previous methods and shows good separation results between the identity and the expression.

\section{Acknowledgments}

The authors would like to thank the development team of the game ``Justice'' for their constructive suggestions and their beautiful 3D character models.

\bibliographystyle{ACM-Reference-Format}
\bibliography{pokerfacegan}

\newpage
\appendix
\onecolumn
\section{Appendix}

\subsection{Details of the differentiable character renderer}
To build our differentiable character renderer, we keep consistent with the game engine -- Unity\footnote{https://unity.com/}, where each vertex of the 3D mesh is associated with four bones. The mesh in the world coordinate $\mathbf{V}^{(q)}_{world}$ can be expressed as follows:
\begin{equation}
\mathbf{V}^{(q)}_{world}  = \sum^{4}_{i=1} {\omega}^{(i)}_{skin} \mathbf{V}^{(q)}\mathbf{M}^{(i)}_{bp} \mathbf{M}^{(i)}_{l2w},
\label{eq:supp_mesh}
\end{equation}
\begin{equation}
\mathbf{M}^{(k)}_{l2w}  = \mathbf{M}^{(r)}_{l2w} \prod^{(k)}_{j=(r)}  \mathbf{M}^{(j)}_{trs}.
\end{equation}
In the above equations, $i$ is the bone ID ($i=1,2,3,4$), $r$ represent the root bone and $k$ is the current bone. $\mathbf{M}_{trs}$ is the homogeneous transformation matrix computed from the bone parameters (can be linearly computed from facial identity parameters $\phi_{idt}$), $\mathbf{M}^{(i)}_{bp}$ is the skinned state of each bone,  $\mathbf{V}^{(q)}$ is the mesh in the local coordinate system, and $\mathbf{M}^{(i)}_{l2w}$ is the local-to-word transformation matrix. Besides, to enrich the rendered characters, we further integrate blend-shapes, a group of expression bases $\mathbf{\delta}$, in our renderer. The Eq (\ref{eq:supp_mesh}), therefore, can be re-formulated as follows:
\begin{equation}
\mathbf{V}^{(q)}_{world}  = \sum^{4}_{i=1} {\omega}^{(i)}_{skin} (\mathbf{V}^{(q)} + \sum_j\phi_{exp, j}\bm{\delta}_j) \mathbf{M}^{(i)}_{bp} \mathbf{M}^{(i)}_{l2w},
\label{eq:supp_blendshape}
\end{equation}
where $\phi_{exp}$ is expression parameters.

{\bf Facial Identity Parameters $\phi_{idt}$.} Here we use the character customization system of the game ``Justice'' as an example to show how the facial identity parameters are configured in our experiments. In Justice, there are 261 facial identity parameters for characters, which are listed in Table~\ref{tab:supp_facial_parameter}. In the column ``Controllers'', the parameters Tx, Ty, Tz, Rx, Ry, Rz, Sx, Sy, Sz correspond to the translation, rotation and scale changes of a facial bone on $x$, $y$ and $z$ axis respectively. The ``\# c'' represents the number of user-adjustable controllers in each group. For those \sout{strikethrough} controllers, their movements are banned considering the symmetry of the human face, which will be set to 0.5 in the output of our method.

{\bf Facial Identity Parameters $\phi_{exp}$.} In this work, we use 22 bases to express common expressions. The basis IDs and their physical meanings are listed in Table~\ref{tab:supp_exp_parameter}.

{\bf Pose parameters $\phi_{pose}$.} Poses are controlled by 3 translation parameters ($-3 \sim 3$) and 3 rotation parameters ($-\pi/2 \sim \pi/2$) on x, y, and z axis respectively. We impose the pose parameters on the mesh before it is fed to the mesh renderer~\cite{genova2018unsupervised}.

\begin{table}[h]
    \centering
    \caption{A detailed configuration of the introduced expression parameters in the game ``Justice''.}
    \begin{tabular}{c|ccccc}
        \toprule
        \textbf{ID} & 0 & 1 & 2 & 3 & 4  \\
        \textbf{Expression} & Eye-close & Upper-lid-raise & Lid-tighten & Inner-brow-raise & Left-outer-brow-raise \\
        \midrule
        \textbf{ID} & 5 & 6 & 7 & 8 & 9 \\
        \textbf{Expression} & Right-outer-brow-raise & Brow-Lower & Jaw-open & Nose-wrinkle & Upper-lip-raise \\
        \midrule
        \textbf{ID} & 10 & 11 & 12 & 13 & 14 \\
        \textbf{Expression} & Down-lip-down & Lip-corner-pull & Left-mouth-press & Right-mouth-press & Lip-pucker \\
        \midrule
        \textbf{ID} & 15 & 16 & 17 & 18 & 19 \\
        \textbf{Expression} & Lip-stretch & Lip-upper-close & Lip-lower-close & Puff & Lip-corner-depress \\
        \midrule
        \textbf{ID} & 20 & 21 &  &   &   \\
        \textbf{Expression} & Jaw-left & Jaw-right & & &  \\
        \bottomrule
    \end{tabular}
    \label{tab:supp_exp_parameter}
\end{table}

\begin{table}[h]
    \centering
    \caption{A detailed configuration of the facial identity parameters in the game ``Justice''.}
    \begin{tabular}{l|l|c}
        \toprule
        \textbf{Bones} & \textbf{Controllers} &  \# c  \\
        \midrule
        eyebrow-head & Tx, Ty, Tz, Rx, \sout{Ry}, Rz, Sx, Sy, Sz &  8 \\
        eyebrow-body & Tx, Ty, Tz, Rx, \sout{Ry}, Rz, Sx, Sy, Sz &  8 \\
        eyebrow-tail & Tx, Ty, Tz, Rx, \sout{Ry}, Rz, Sx, Sy, Sz &  8 \\
        \midrule
        eye & Tx, Ty, Tz, Rx, Ry, Rz, \sout{Sx}, \sout{Sy}, \sout{Sz} &  6 \\
        outside eyelid & Tx, Ty, Tz, Rx, Ry, Rz, Sx, Sy, Sz & 9 \\
        inside eyelid & Tx, Ty, Tz, Rx, Ry, Rz, Sx, Sy, Sz & 9 \\
        lower eyelid & Tx, Ty, Tz, Rx, Ry, Rz, Sx, Sy, Sz &  9 \\
        inner eye corner & Tx, Ty, Tz, Rx, Ry, Rz, Sx, Sy, Sz &  9 \\
        outer eye corner & Tx, Ty, Tz, Rx, Ry, Rz, Sx, Sy, Sz &  9 \\
        \midrule
        nose body & \sout{Tx}, Ty, Tz, Rx, \sout{Ry}, \sout{Rz}, \sout{Sx}, \sout{Sy}, \sout{Sz} &  3 \\
        nose bridge & \sout{Tx}, Ty, Tz, Rx, \sout{Ry}, \sout{Rz}, Sx, Sy, Sz &  6 \\
         nose wing & Tx, Ty, Tz, Rx, Ry, Rz, Sx, Sy, Sz &  9 \\
         nose tip & \sout{Tx}, Ty, Tz, Rx, \sout{Ry}, \sout{Rz}, Sx, Sy, Sz &  6 \\
         nose bottom  & \sout{Tx}, Ty, Tz, Rx, \sout{Ry}, \sout{Rz}, Sx, Sy, Sz &  6 \\
        \midrule
        mouth & \sout{Tx}, Ty, Tz, Rx, \sout{Ry}, \sout{Rz}, \sout{Sx}, \sout{Sy}, \sout{Sz} &  3 \\
        middle upper lip & \sout{Tx}, Ty, Tz, Rx, \sout{Ry}, \sout{Rz}, Sx, Sy, Sz &  6 \\
        outer upper lip & Tx, Ty, Tz, Rx, Ry, Rz, Sx, Sy, Sz &  9 \\
        middle lower lip & \sout{Tx}, Ty, Tz, Rx, \sout{Ry}, \sout{Rz}, Sx, Sy, Sz &  6 \\
        outer lower lip & Tx, Ty, Tz, Rx, Ry, Rz, Sx, Sy, Sz &  9 \\
        mouth corner & Tx, Ty, Tz, Rx, Ry, Rz, Sx, Sy, Sz &  9 \\
        \midrule
        forehead & \sout{Tx}, Ty, Tz, Rx, \sout{Ry}, \sout{Rz}, Sx, Sy, Sz &  6 \\
        glabellum & \sout{Tx}, Ty, Tz, Rx, \sout{Ry}, \sout{Rz}, Sx, Sy, Sz &  6 \\
        cheekbone & Tx, Ty, Tz, Rx, Ry, \sout{Rz}, \sout{Sx}, \sout{Sy}, \sout{Sz} &  5 \\
        risorius &  Tx, Ty, Tz, Rx, Ry, \sout{Rz}, \sout{Sx}, \sout{Sy}, \sout{Sz} &  5 \\
        cheek &  Tx, Ty, Tz, Rx, Ry, Rz, \sout{Sx}, \sout{Sy}, \sout{Sz} &  6 \\   
        jaw & \sout{Tx}, Ty, Tz, Rx, \sout{Ry}, \sout{Rz}, Sx, Sy, Sz &  6 \\
        lower jaw & Tx, Ty, Tz, Rx, Ry, Rz, Sx, Sy, Sz &  9 \\
        mandibular & Tx, Ty, Tz, Rx, Ry, Rz, Sx, Sy, Sz &  9 \\
        outer jaw & Tx, Ty, Tz, Rx, Ry, Rz, Sx, Sy, Sz &  9 \\
        \bottomrule
    \end{tabular}
    \label{tab:supp_facial_parameter}
\end{table}

\clearpage
\subsection{More comparison results}
\begin{figure*}[ht]
\centering
  \includegraphics[width=0.9\linewidth]{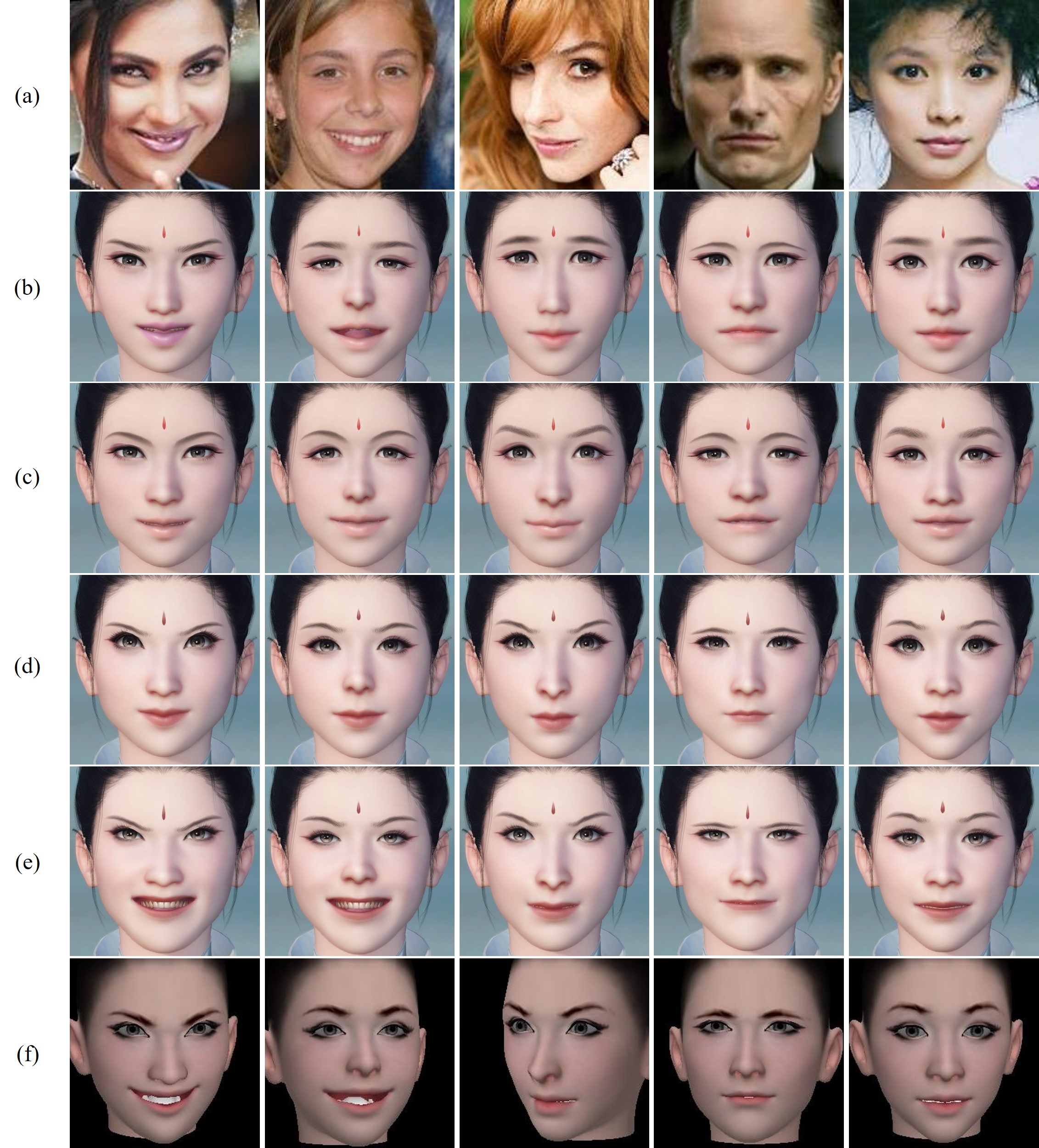}
  \caption{More visual comparisons of different methods for in-game character auto-creation. (a) Aligned input photos, (b) The results of F2P~\cite{Shi_2019_ICCV}, (c) The results of FR-F2P~\cite{shi2020fast}, (d) Our results (identity-only) (e) Our results (with both identity and expression) (f) Our results rendered by our differentiable character renderer. (Samples are from CelebA test set~\cite{liu2015faceattributes})}
  \label{fig:big-supp}
\end{figure*}

\newpage
\subsection{More disentanglement results}

\begin{figure*}[ht]
\centering
  \includegraphics[width=0.9\linewidth]{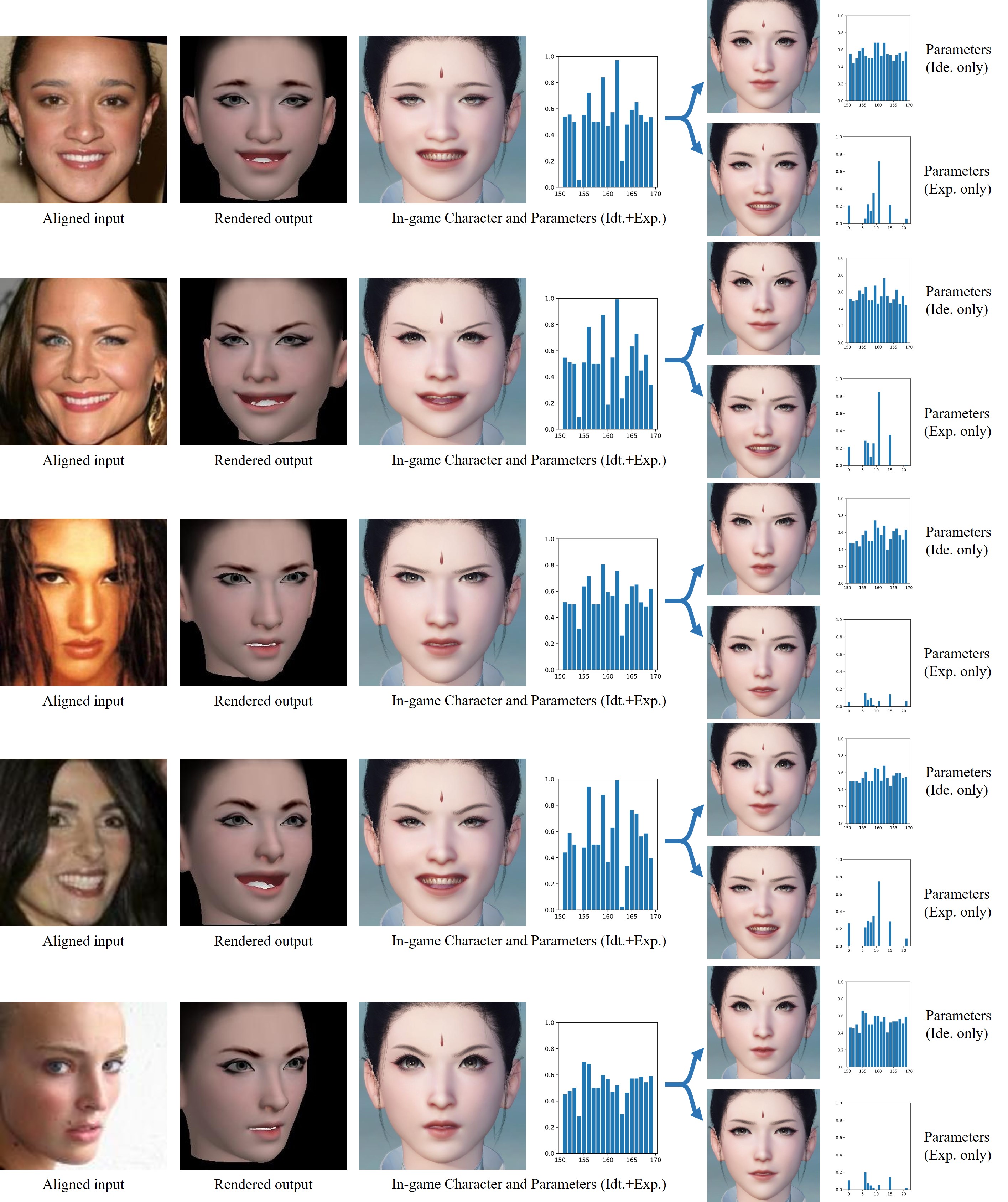}
  \caption{More examples of the separation results of the identity (Idt.) and expression (Exp.) by using our method. Notice that there are in total of 261 facial identity parameters for each face, for better visualization, we only plot its 151 $\sim$ 170 dimensions (the mouth part). (Samples are from CelebA test set~\cite{liu2015faceattributes})}
  \label{fig:details-supp}
\end{figure*}

\end{document}